\documentclass{article}
\def\mc{\mathcal}

\usepackage{ijcai23}

\usepackage{times}
\usepackage{soul}
\usepackage{url}
\usepackage[hidelinks]{hyperref}
\usepackage[utf8]{inputenc}
\usepackage[small]{caption}
\usepackage{graphicx}
\usepackage{amsmath}
\usepackage{amsthm}
\usepackage{amsfonts}
\usepackage{booktabs}
\usepackage{multirow}
\usepackage{algorithm}
\usepackage{algorithmic}
\usepackage[switch]{lineno}

\usepackage{bbold}
\usepackage{mathtools}

\title{Calibrating a Deep Neural Network with Its Predecessors}

\author{
    Linwei Tao$^1$\and
    Minjing Dong$^1$\and
    Daochang Liu$^{1}$\and
    Changming Sun$^{2}$\And
    Chang Xu$^{1}$
    \affiliations
    $^1$School of Computer Science, Faculty of Engineering, University of Sydney, Australia\\
    $^2$CSIRO’s Data61, Australia
    \emails
    \{linwei.tao, mdon0736, daochang.liu\}@sydney.edu.au, 
    changming.sun@csiro.au,
    c.xu@sydney.edu.au
}

\begin{document}

\maketitle

\begin{abstract}
    Confidence calibration - the process to calibrate the output probability distribution of neural networks - is essential for safety-critical applications of such networks. Recent works verify the link between mis-calibration and overfitting. However, early stopping, as a well-known technique to mitigate overfitting, fails to calibrate networks. In this work, we study the limitions of early stopping and comprehensively analyze the overfitting problem of a network considering each individual block. We then propose a novel regularization method, predecessor combination search (PCS), to improve calibration by searching a combination of best-fitting block predecessors, where block predecessors are the corresponding network blocks with weight parameters from earlier training stages. PCS achieves the state-of-the-art calibration performance on multiple datasets and architectures. In addition, PCS improves model robustness under dataset distribution shift. Supplementary material and code are available at https://github.com/Linwei94/PCS
\end{abstract}

\section{Introduction}

Deep neural networks (DNNs) have achieved great successes across a variety of domains, especially on classification related tasks such as object detection~\cite{wang2022yolov7,zhou2021probabilistic,qiao2021detectors} and image classification~\cite{pham2021meta,wortsman2022model}, reaching prediction accuracy far beyond human beings. However, they still suffer from mis-calibrated predictions in the sense that the prediction probability cannot represent the ground-truth probability.This may lead to fatal problems when any safety-critical downstream tasks such as autonomous driving~\cite{bojarski2016end} and medical diagnosis~\cite{caruana2015intelligible} rely heavily on the prediction probability.

The underlying cause for mis-calibrated predictions is associated with the capacity of modern neural networks that makes them vulnerable to overfitting~\cite{guo2017calibration}. ~\cite{mukhoti2020calibrating} show that overfitting in modern neural networks mostly results from the overconfidence on mis-classified samples and they empirically verify the strong connection between the overfitting issue and calibration performance. Given this observation, some regularization techniques such as weight decay~\cite{guo2017calibration}, label smoothing~\cite{muller2019does}, and data augmentation~\cite{thulasidasan2019mixup,hendrycks2019augmix} are introduced to improve model calibration. 

Early stopping~\cite{prechelt1998early} is another well-known regularization method, which suspends training once the model performance stops improving on a hold out validation dataset. ~\cite{mukhoti2020calibrating} conduct a series of empirical experiments and demonstrate that early stopping on training according to multiple criteria fails to yield a well-calibrated model. We mainly attribute this sub-optimal solution to the unitary strategy of conventional early stopping techniques which treat the entire network as a whole. Specifically, an early stopping technique takes a DNN as a black box without investigating the internal components, i.e., the blocks inside the network. However, the increasing depth of modern DNNs makes the optimization more challenging, which could lead to discrepancies of convergence speeds of different blocks in DNNs. Thus, any model calibration via a conventional early-stopping technique could be a sub-optimal solution.

In this paper, instead of taking a DNN as a whole, we consider a block in a DNN as the basic unit and explore the overfitting problem in each block. We empirically observe that blocks in a network overfit at different stages during training. Unlike the conventional early stopping approach~\cite{prechelt1998early} that stops the training of the whole network at a certain point to form a network predecessor, we propose to stop the training of each block at its own best-fitting block predecessor to improve model calibration. However, the blocks in DNNs are strongly coupled with each other, and the early stopping  of an individual block independently does not ensure an optimal solution.
To achieve an effective and adaptive early stopping for each block, we take into consideration all possible block predecessor combinations. Our objective is to discover the predecessor combination (PC) with better calibration performance.
We propose a neural architecture search inspired approach, predecessor combination search (PCS) to calibrate the DNNs, which performs a differential search of the optimal block predecessor combination through a relaxation of the search space as well as a predecessors evaluation estimator.

Our contribution can be summarized as follows:
\textbf{(1)} We study the overfitting problem of individual blocks empirically and show that different blocks reach their own overfitting points at different stages of training.
\textbf{(2)} We propose a novel differential PCS method to search a better-calibrated model together with a sampling strategy to improve searching efficiency.
\textbf{(3)} PCS achieves state-of-the-art results for both pre- and post-temperature scaling~\cite{guo2017calibration} on a variety of datasets and architectures via a large number of experiments. We show that PCS works well on out-of-distribution (OoD) samples by shifting the dataset from CIFAR-10 to SVHN~\cite{goodfellow2013multi} and CIFAR-10-C~\cite{hendrycks2018benchmarking}.

\section{Related Works}
Due to the mis-calibration problem in modern neural networks~\cite{guo2017calibration} and the significant importance of calibration, many techniqueshave been proposed in recent years. The current calibration methods could be divided into three categories. The first category modifies the training loss by replacing the conventionally used cross-entropy loss with a mean square error loss~\cite{hui2020evaluation} or a focal loss~\cite{gupta2020calibration} or by adding an auxiliary regularization loss such as the MMCE loss~\cite{kumar2018trainable} and the AvUC loss~\cite{krishnan2020improving}. ~\cite{bohdal2021meta} propose a differentiable surrogate for expected calibration error that improves the calibration performance directly. The recent work in ~\cite{karandikar2021soft}
introduces a differentiable bin membership function and applies it on bin-based metrics such as expected calibration error (ECE)~\cite{guo2017calibration} to make it become a differentiable auxiliary calibration loss.
 
Another category is the post-hoc calibration approaches that improve the calibration performance by modifying the prediction logits. Platt scaling~\cite{platt1999probabilistic} learns parameters to perform a linear transformation on the original prediction logits. Isotonic regression~\cite{zadrozny2002transforming} learns piece-wise functions to transform the original prediction logits. Histogram binning~\cite{zadrozny2001obtaining} obtains calibrated probability estimates from decision trees and naive Bayesian classifiers. Bayesian binning into quantiles (BBQ)~\cite{naeini2015obtaining} is an extension of histogram binning with Bayesian model averaging. Beta calibration~\cite{kull2017beta} is proposed for binary classification and~\cite{kull2019beyond} generalize the beta calibration method from binary classification to multi-classification with Dirichlet distributions.
~\cite{wenger2020non} employ a non-parametric representation using a latent Gaussian process. Among these methods, temperature scaling is the most popular post-hoc approach, which tunes the temperature parameter of the softmax function that minimizes the negative log likelihood (NLL) and does not change the prediction results. In this work, we present the calibration performance with both before and after temperature scaling. 
 
All other regularization methods that can calibrate networks form the third category. Label smoothing~\cite{muller2019does} implicitly calibrates networks by artificially softening targets to prevent a model from overfitting to the ``hard label". Mixup~\cite{thulasidasan2019mixup} and AugMix~\cite{hendrycks2019augmix} are two popular data augmentation techniques for calibration. The mix step in data augmentation increases the generality of datasets and reduces the influence of hard samples that can easily cause over-confidence problem. Weight decay, which dominated regularization methods for neural networks in the past, is now less often used by modern neural networks. However, it still plays an important role in improving model calibration~\cite{guo2017calibration}. Learning with Retrospection(LWR)~\cite{deng2021learning} makes use of the learned information in the past epochs to guide the subsequent training, which benefit the classification prediction and uncertainty.

\section{Problem Formulation}
Considering a dataset $\mathcal{D} = \langle(x_i, y_i)\rangle_{i=1}^N$ with $N$ samples from a joint distribution $(\mathcal{X}, \mathcal{Y})$, the ground-truth class label is $y_i \in \{1, 2, ..., \mathcal{K}\}$, where $\mathcal{K}$ denotes the number of classes. The probability for a class $y_i$ on a given input $x_i$ predicted by network $F$ with model parameters $\Theta$ is denoted as $\hat{p}_{i,y_i} = F_\Theta(y_i|x_i)$. The predicted label $\hat{y}_i$ and the corresponding confidence $\hat{p}_i$ are defined as
\begin{align}
    &\hat{y}_i = \mathrm{argmax}_{y_i \in \{1, 2, ..., \mathcal{K}\}} \; \hat{p}_{i,y_i}, \; \notag\\
    &\hat{p}_i = \mathrm{max}_{y_i \in \{1, 2, ..., \mathcal{K}\}} \; \hat{p}_{i,y_i}.
\end{align}
When the model is \emph{perfectly calibrated}, the prediction confidence $\hat{p}$ is expected to represent the real probability $p$ for each sample $x_i$ with class label $y_i$. In other words, the model accuracy $\mathbb{P}(\hat{y} = y | \hat{p} = p)$ is $p$, for all $p\in[0,1]$.

ECE is a widely-accepted metric to measure calibration performance. Formally, the ECE is defined as the expected absolute difference between the model's confidence and its accuracy, which can be formulated as

\begin{equation}
\label{eq:ECE}
    \mathrm{ECE} = \mathbb{E}_{\hat{p}} \big[ \left| \mathbb{P}(\hat{y} = y | \hat{p}) - \hat{p} \right|  \big].
\end{equation}

Due to the finite samples in datasets, ~\cite{guo2017calibration} estimate ECE by dividing confidence $p\in[0,1]$ into $\mathbb{B}$ equal-width bins. $B_i$ denotes the set of samples with confidences within $\left(\frac{i-1}{\mathbb{B}}, \frac{i}{\mathbb{B}} \right]$. Let $I_i$ and $C_i$ denote the accuracy and average confidence of all samples in bin $B_i$ respectively. Accuracy of bin $B_i$ is computed as \(I_i = \frac{1}{|B_i|} \sum_{j \in B_i} \mathbb{1} \left(\hat{y}_j = y_j\right) \), where $\mathbb{1}$ is the indicator function, and $\hat{y}_j$ and $y_j$ are the predicted and ground-truth labels for the $j^{\mathrm{th}}$ sample. Similarly, the confidence $C_i$ of the $i^{\mathrm{th}}$ bin is computed as \(C_i = \frac{1}{|B_i|} \sum_{j \in B_i} \hat{p}_j \), i.e.,\ $C_i$ is the average confidence of all samples in the bin. 
Thus, in practice, ECE is formulated as the weighted average of accuracy-confidence difference for the bins:
\begin{equation}
\label{eq:ECE_appro}
    \mathrm{ECE} = \sum_{i=1}^{\mathbb{B}} \frac{|B_i|}{N} \left| I_i - C_i \right|.
\end{equation}

Along with ECE, the maximum calibration error (MCE) is proposed to minimize the influence of worst-case confidence deviation, which is defined as the maximum difference of bins' accuracy and confidence: $\mathrm{MCE} = \max_{i\in 1,\dots,\mathbb{B}} \left| I_i - C_i \right|$.

\begin{figure*}[t]
	\centering
	\includegraphics[width=\linewidth]{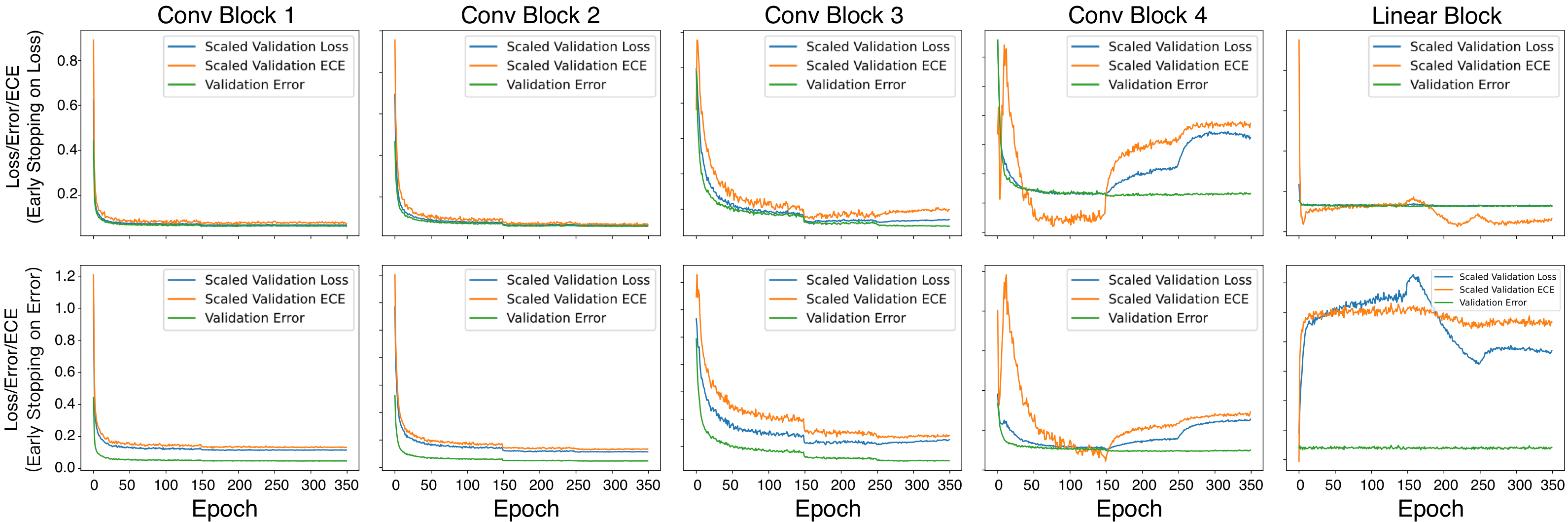}
	\caption{
		\textbf{Empirical evidence on that different blocks have different overfitting behaviors.} A ResNet-50 on CIFAR-10 is first trained with 350 epochs with the cross-entropy loss and learning rate scheduling at epochs 150 and 250. ResNet-50 has 5 blocks, i.e., 4 convolutional blocks and 1 final linear layer. Each sub-figure shows the overfitting issue of block $f_i$ through the training epochs with $\tilde{F}_i$ fixed at potential ``sweet point" predecessors. Top row and bottom row represent $\tilde{F}_i$ fixed at different predecessors, which are the ``sweet point" epoch of the whole model by early stopping on loss (epoch 151) and error (epoch 281) respectively. Columns represent different block $f_i$ under study. All combinations are fine-tuned with one epoch on the training set. 
	}
	\label{fig:motivation}
\end{figure*}

~\cite{guo2017calibration} and~\cite{mukhoti2020calibrating} state that the mis-calibration problem of networks is strongly related to overfitting on training sets.
Early stopping, as a well-known regularization technique, mitigates the overfitting problem by stopping the whole network at the best-fitting epoch. According to previous observations, early stopping should yield a well calibrated model. However, the calibration performance of an early stopped model is often far from satisfactory. Thus, we hypothesize that shallow and deep blocks may suffer from different degrees of overfitting during optimization and a unified regularization criteria to the entire network $F_\Theta$ could lead to a sub-optimal solution for model calibration. Instead, it would be more desirable to resolve the overfitting problems block-wise and explore an adaptive regularization approach to improve the model calibration performance. 

\subsection{Overfitting in Blocks}
Most advanced CNN model architectures are the stack of blocks. Suppose there are $M$ blocks in the network, the model parameters can be represented as $\Theta=\{\theta_i|i=1,2,\dots,M\}$, where $\theta_i$ denotes the parameter of the $i^{\textrm{th}}$ block. $\Theta^j$ and $\theta^j$ are used to identify the network and block parameters at the $j^{\textrm{th}}$ training epoch. 
The degree of fitting of $F_\Theta$ is represented by the trend of validation NLL loss $\mathcal{L}_{NLL}(F_{\Theta^j}(x), y), j=1,2,...,T_{train}$, where $T_{train}$ is the number of total training epochs over the training process. Conventional early stopping aims to find the weights at a ``sweet point"\footnote{A sweet point is defined as the balance point between underfitting and overfitting, which normally refers to the lowest point of the validation loss curve.} epoch to balance between underfitting and overfitting. However, there is no definition of overfitting of individual network blocks.
In order to find the ``sweet point" epoch of each block, we need to measure the degree of overfitting of an individual block $f_i$. We formally define the \emph{\textbf{predecessors}} of block $f_i$ as \{$f_i^{j}| j\in \{1,2,\dots,T_{train}\}\}$, where $f_i^{j}$ denotes the $i^{\textrm{th}}$ block with weight at the $j^{\textrm{th}}$ epoch. 
To achieve the overfitting measurement of $f_i$, we treat the other blocks of the network $\tilde{F}_i=\{f_{i^-}|i^-\in \{1,2,\dots,M\}\setminus\{i\}\}$ as constant mappings.
Then the overfitting degree of block $f_i$ could be represented by the trend of validation NLL loss over the training process $\mathcal{L}_{NLL}(F(x), y;f_i^j), j=1,2,...,T_{train}$, with the weights of other network blocks $\tilde{F}_i$ fixed. 
Based on this idea, an empirical study is conducted to investigate the overfitting behaviors of individual network blocks.

We present the empirical study in Figure~\ref{fig:motivation}, demonstrating the different overfitting behaviors of individual network blocks. A ResNet-50 is first trained on CIFAR-10 for $T_{train}=350$ epochs.  
Then each sub-figure shows the evaluation of overfitting behavior of block $f_i$ and each block other than $f_i$ is fixed to a certain predecessor. Note that it is non-trivial to properly select the fixed predecessors for all other blocks $\tilde{F}_i$ and ideally these blocks need to be at their own ``sweet points".
We adopt a simplified approach and fix $\tilde{F}_i$ at the ``sweet point" of the whole model as an approximation, i.e., $\tilde{F}_i=\{f_{i^-}^{\rho^*}|i^-\in \{1,2,\dots,M\}\setminus\{i\}\}$, where $\rho^*$ is obtained either by early stopping the whole model based on the validation loss or classification error.
Apart from the validation loss, the validation ECE and classification error are also plotted in Figure~\ref{fig:motivation}. 
The proportionally scaled values of the validation loss and ECE are reported for better visualisation.
Through the variation of the validation loss of each block $f_i$ in this experiment as shown in Figure~\ref{fig:motivation}, we have the following intriguing observations on overfitting in blocks.

\noindent
\textbf{1.} \textbf{Deeper convolutional blocks tend to have more severe overfitting problems with training.} From the first four columns in Figure~\ref{fig:motivation}, we observe that all convolutional blocks with the same block index share a similar overfitting pattern for both settings. However, when we look into each row, a different pattern shows on the convolutional blocks compared with each other. The deeper convolutional block, i.e., $f_4$, starts rapid overfitting after the first learning rate scheduler point.

\noindent
\textbf{2.} \textbf{The overfitting problem of the final linear layer is much more complex and depends heavily on other blocks.} When it comes to the final linear block $f_5$ at the last column, the loss presents a totally reverse pattern with different $\tilde{F}_i$. The final block $f_5$ with $\tilde{F}_i$ fixed at predecessors early stopping on loss has a slight overfitting problem from epoch $50$ to the first learning rate scheduler point at epoch $150$, while $f_5$ with $\tilde{F}_i$ fixed at predecessors early stopping on classification error continues to overfit from the very first epoch until a few epochs after the first learning rate scheduler point while the error remains at the same level. 

\noindent
\textbf{3.} \textbf{Mis-calibration is linked with block overfitting.} When we look at the variation of ECE and validation loss, we observe that the overfitting trend of an individual block is consistent with the variation of ECE, which further extends the link between mis-calibration and overfitting at block-wise level.

According to these observations, it could be very difficult to identify the ideal PC due to the inter-dependency among blocks (the linear blocks in the two rows have totally different behaviours). In other words, simply finding the best-fitting blocks individually and combining them together without taking other blocks into account may not produce a best-fitting model. 
This motivates us to explore an automatic searching algorithm to find a group of predecessors  at ``sweet points" that allows $F^*= \{f_i^{\rho_i^*}|i=1,2,\dots,M\}$ to best-fit to the training set. Here,  $\rho_i^*$ denotes the optimal block predecessor choice of block $f_i$.

\section{Methodology}
\label{Methodology}
Based on the observation on Figure~\ref{fig:motivation}, we propose to explore a PC representation $\mc{P}=\{\rho_i|i=1,2,\dots,M\}$ to tackle the overfitting issue in blocks and improve model calibration performance. The objective can be formulated as
\begin{equation} \label{eq:objective1}
    \operatorname*{min}_{\mc{P}}
    \left(ERR(F) + \lambda \cdot ECE(F)\right)
\end{equation}
where $F$ is $\{f_i^{\rho_i}|i=1,2,\dots,M\}$, $\lambda$ denotes a hyperparameter, $\rho_i\in \{1,2,\dots,T_{train}\}$ indicates the predecessor selection of block $f_i$, $ERR$ and $ECE$ are the classification error and ECE respectively. The direct optimization of Eq.~(\ref{eq:objective1}) is not feasible via gradient descent due to two issues. First, the discrete representation $\mc{P}$ makes Eq.~(\ref{eq:objective1}) an optimization problem over a discrete domain. Second, both terms $ERR$ and $ECE$ in our objective are not differentiable. Thus, we introduce a PCS framework to tackle the aforementioned issues. Specifically, we first introduce differentiable combination sampling through a continuous relaxation of the PC representation. Proxy classification error and ECE landscape are then introduced to achieve differential optimization of our objective through an estimator for predicting the classification error and ECE.

\subsection{Differentiable Combination Sampling}
In our PCS framework, the objective is to discover the optimal selection of predecessors from candidate sets for each block. 
To learn the selection, we first exploit a $K$-dimensional trainable parameter $\alpha_i \in \mathbb{R}^K$ for this predecessor selection, where $K$ denotes the number of candidates.
And $\rho_i$ can be obtained by the argmax of the selection parameter $\alpha_i$ and further represented in a one-hot encoding format $\rho_i\in \mathbb{R}^K$.
The PC representation $\mc{P}$ can be written as:

\begin{align}
\label{eq:weight_combination1}
&\mc{P}=\{\rho_i | i=1,2, \dots, M\}, \notag\\
\text{s.t.}\quad &\rho_i=\operatorname*{one-hot}(\operatorname*{argmax} \alpha_i).   
\end{align}

To relax the discrete PC representation for gradient-based optimization, we use the Gumbel-Softmax trick~\cite{jang2016categorical} to approximate the one-hot distribution and introduce randomness. The $\rho_i$ in Eq.~(\ref{eq:weight_combination1}) can be relaxed as:

\begin{equation}
\label{eq:weight_combination2}
   \tilde{\rho}_i^k= \frac{\exp((\alpha_i^k+\xi_i^k)/\tau)}{\sum_{k'=1}^{K}\exp((\alpha_i^{k'}+\xi_i^{k'})/\tau)},
\end{equation}
where $\tau$ is the temperature parameter, $\xi_i^k$ is an i.i.d sample from Gumbel(0, 1), $k$ and $k'$ denote the $k^{\textrm{th}}$ and $k'^{\textrm{th}}$ logit of corresponding $K$-dimensional vector respectively. We denote the relaxed PC representation as $\tilde{\mc{P}}=\{\tilde{\rho}_i | i=1,2, \dots, M\}$. With the learning of $\alpha_i$, $\tilde{\mc{P}}$ explores the random combination at the beginning of search and gradually converges to a relatively stable state. 

For simplicity, we denote the model with $\tilde{\mc{P}}$ as $F_{\tilde{\mc{P}}}$. The classification error and ECE on the validation set can be denoted as $ERR(\mathcal{D}_{val}, F_{\tilde{\mc{P}}})$ and $ECE(\mathcal{D}_{val}, F_{\tilde{\mc{P}}})$  respectively. Since the combination is hard-combined, we fine-tune $F_{\tilde{\mc{P}}}$ with one more epoch on $\mathcal{D}_{train}$, denoted as $F_{\tilde{\mc{P}}}^*$. In PCS, the original objective (Eq.~(\ref{eq:objective1})) can be reformulated as:
\begin{align}
\label{eq:objective2}
\operatorname*{min}_A \left(ERR(\mathcal{D}_{val}, F_{\tilde{\mc{P}}}^*) + \lambda  \cdot ECE(\mathcal{D}_{val}, F_{\tilde{\mc{P}}}^*)\right),
\end{align}
where $A=\{\alpha_i | i=1,2, \dots, M\}$ is the collection of learnable selection parameters for all blocks.

\begin{table*}[!t]
	\centering
	\scriptsize
	\resizebox{\linewidth}{!}{%
		\begin{tabular}{cccccccccccccccc}
			\toprule
			\textbf{Dataset} & \textbf{Model} & \multicolumn{2}{c}{\textbf{Weight Decay}} &
			\multicolumn{2}{c}{\textbf{Brier Loss}} & \multicolumn{2}{c}{\textbf{MMCE}} &
			\multicolumn{2}{c}{\textbf{Label Smoothing}} & \multicolumn{2}{c}{\textbf{FL-3}} &
			\multicolumn{2}{c}{\textbf{FLSD-53}} &
			\multicolumn{2}{c}{\textbf{PCS}} \\
			\textbf{} & \textbf{} &
			\multicolumn{2}{c}{\cite{guo2017calibration}} &
			\multicolumn{2}{c}{\cite{brier1950verification}} & \multicolumn{2}{c}{\cite{kumar2018trainable}} &
			\multicolumn{2}{c}{\cite{szegedy2016rethinking}} & \multicolumn{2}{c}{\cite{mukhoti2020calibrating}} &
			\multicolumn{2}{c}{\cite{mukhoti2020calibrating}} &
			\multicolumn{2}{c}{Ours} \\
			&& Pre T & Post T & Pre T & Post T & Pre T & Post T & Pre T & Post T & Pre T & 
			Post T & Pre T & Post T & Pre T & Post T \\
			\midrule
			
			\multirow{4}{*}{CIFAR-100} & ResNet-50&17.52&3.42(2.1)&6.52&3.64(1.1)&15.32&2.38(1.8)&7.81&4.01(1.1)&5.13&1.97(1.1)&4.5&2.0(1.1)&\textbf{2.0}&\textbf{2.0(1.0)}\\
			& ResNet-110&19.05&4.43(2.3)&7.88&4.65(1.2)&19.14&3.86(2.3)&11.02&5.89(1.1)&8.64&3.95(1.2)&8.56&4.12(1.2)&\textbf{1.76}&\textbf{1.76(1.0)}\\
			& Wide-ResNet-26-10&15.33&2.88(2.2)&4.31&2.7(1.1)&13.17&4.37(1.9)&4.84&4.84(1.0)&2.13&2.13(1.0)&3.03&1.64(1.1)&\textbf{1.92}&\textbf{1.55(1.1)}\\
			& DenseNet-121&20.98&4.27(2.3)&5.17&2.29(1.1)&19.13&3.06(2.1)&12.89&7.52(1.2)&4.15&1.25(1.1)&3.73&1.31(1.1)&\textbf{2.75}&\textbf{1.18(1.1)}\\
			\midrule
			\multirow{4}{*}{CIFAR-10} & ResNet-50&4.35&1.35(2.5)&1.82&1.08(1.1)&4.56&1.19(2.6)&2.96&1.67(0.9)&1.48&1.42(1.1)&1.55&0.95(1.1)&\textbf{0.80}&\textbf{0.53(1.1)}\\
			& ResNet-110&4.41&1.09(2.8)&2.56&1.25(1.2)&5.08&1.42(2.8)&2.09&2.09(1.0)&1.55&1.02(1.1)&1.87&1.07(1.1)&\textbf{0.57}&\textbf{0.57(1.0)}\\
			& Wide-ResNet-26-10&3.23&0.92(2.2)&1.25&1.25(1.0)&3.29&0.86(2.2)&4.26&1.84(0.8)&1.69&0.97(0.9)&1.56&0.84(0.9)&\textbf{0.99}&\textbf{0.43(1.2)}\\
			& DenseNet-121&4.52&1.31(2.4)&1.53&1.53(1.0)&5.1&1.61(2.5)&1.88&1.82(0.9)&1.32&1.26(0.9)&1.22&1.22(1.0)&\textbf{0.78}&\textbf{0.78(1.0)}\\
			\midrule
			Tiny-ImageNet & ResNet-50&15.32&5.48(1.4)&4.44&4.13(0.9)&13.01&5.55(1.3)&15.23&6.51(0.7)&1.87&1.87(1.0)&1.76&1.76(1.0)&\textbf{1.32}&\textbf{1.32(1.0)}\\
			\bottomrule
		\end{tabular}%
	}
	\caption{\textbf{Calibration Performance.}\quad  ECE $(\%)$, being the lower the better, is evaluated for different methods. Both pre and post temperature scaling (Pre T and Post T in Table) results are reported. The optimal temperature is obtained on the validation set and is included in brackets.}
	\label{table:ece_tab1}
\end{table*}

\subsection{Proxy Classification Error and ECE Landscape}
For differential optimization of PC representation $\tilde{\mc{P}}$, we use a trainable estimator $\psi$ to obtain proxies $\hat{ERR}, \hat{ECE} = \psi(\tilde{\mc{P}})$ to approximate the classification error and ECE. 
Since the input $\tilde{\mc{P}}$ is a sequential data, we utilize a one-layer long short-term memory (LSTM)  to build the estimator $\psi$: $\mathbb{R}^{M\times K} \rightarrow \mathbb{R}^d$ mapping $\tilde{\mc{P}}$ to a $d$-dimensional embedding vector and a linear layer: $\mathbb{R}^{d} \rightarrow \mathbb{R}^2$ outputting $\hat{ERR}$ and $\hat{ECE}$.
The estimator $\psi$ is trained with a weighted mean squared error loss function:

\begin{align}
\label{eq:estimator1}
\operatorname*{min}_\psi L(\psi)\notag
=&\frac{1}{T_{se}}\sum_{t=1}^{T_{se}} (\hat{ERR}^{(t)}-ERR^{(t)}(\mathcal{D}_{val}, F_{\tilde{\mc{P}}}^*))^2\notag\\ +&\gamma(\hat{ECE}^{(t)}-ECE^{(t)}(\mathcal{D}_{val}, F_{\tilde{\mc{P}}}^*))^2,
\end{align}

\noindent where $\gamma$ is a hyperparameter to control the loss ratio of ECE, $T_{se}$ is the total searching steps and the superscript ${(t)}$ indicates the evaluation results at the $t^{\textrm{th}}$ time step. All pairs of $\tilde{\mc{P}}$ and its corresponding classification error and ECE are stored in a memory $\Pi$ to optimize estimator $\psi$. After each searching step $t$, memory $\Pi$ is updated by $\Pi=\Pi\cup\{(\tilde{\mc{P}}^{(t)}:(ECE^{(t)}, ERR^{(t)}))\}$. 
We can then use the optimized estimator $\psi^*$ to reformulate PCS objective Eq.~(\ref{eq:objective2}):
\begin{equation}
\begin{aligned}
\label{eq:objective3}
&\operatorname*{min}_A (\hat{ERR}^*  + \lambda \hat{ECE}^*),\\
& \text{where} \; \hat{ERR}^*, \hat{ECE}^* = \psi^{*}(\tilde{\mc{P}}).\\
\end{aligned}
\end{equation}
The gradients of $\hat{ERR}^*$ and $\hat{ECE}^*$ can be used to optimize $\tilde{\mc{P}}$ and thus $A$:
\begin{align}
\label{eq:updateA}
A'\leftarrow A-\eta\cdot\nabla_A(\hat{ERR}^* + \lambda \hat{ECE}^*),
\end{align}
\noindent where $A'$ is the new predecessor selection parameter and $\eta$ is the learning rate. At the next searching time step, the corresponding $\tilde{\mc{P}}'$ is based on $A'$, and memory $\Pi$ is updated to $\Pi=\Pi\cup\{(\tilde{\mc{P}^{'}}:(ECE^{'}, ERR^{'})\}$.

\noindent
\textbf{Remark: Search Procedure}
\quad 
We first train model $F_\Theta$ with $T_{train}$ epochs and randomly store weight parameters at $K$ different epochs.  After that, we randomly initialize a warm-up population $\mathbb{H}$ of size $\mc{S}$ and evaluate the classification error $ERR^{(t)}(\mathcal{D}_{val}, F_{\tilde{\mc{P}}_i}^*)$ and ECE $ECE^{(t)}(\mathcal{D}_{val}, F_{\tilde{\mc{P}}_i}^*)$ of $\tilde{\mc{P}}_i \in \mathbb{H}$. The combination-performance pairs are then stored to memory $\Pi$, which is used to warm up estimator $\psi$ and equip $\psi$ with prior knowledge about classification error and ECE before searching.
After the initial training of model and warming up of $\psi$, the searching procedure is conducted with $T_{se}$ steps. In each  step, one PC representation $\tilde{\mc{P}}^{(t)}$ is sampled based on current $A^{(t)}$. The corresponding $F_{\tilde{\mc{P}}^{(t)}}$ is then fine-tuned with one epoch and evaluated to obtain a validation error and ECE for the training of $\psi$. $A^{(t)}$ is then optimized with the proxy classification error $\hat{ERR}$ and proxy ECE $ \hat{ECE}$. 

\noindent
\textbf{Remark: Search Space Sampling Strategy}
\label{Sampling_Strategy}
\quad Each selection $\rho$ is an integer between $1$ and training epoch $T_{train}$, and thus the candidate size $K$ is $T_{train}$. However, storing weight parameters over all training epochs can be very storage-intensive. For instance, storing all candidates of a ResNet-50 training with 350 epochs can take up to 33GB. Considering that the model parameters of adjacent epochs are similar, especially at late training stages, we propose four sampling strategies to reduce the size of search space and improve storage efficiency, which are 
(1)  \textit{Random Sampling}, randomly sampling ${K}$ different epochs ranging from $1$ to $T_{train}$;
(2)  \textit{Uniform Sampling}, uniformly sampling ${K}$ epochs ranging from $1$ to $T_{train}$;
(3)  \textit{Laplace Sampling}, due to model parameters changing much faster at earlier epochs, sampling ${K}$ epochs ranging from $1$ to $T_{train}$ with a Laplace distribution centering at 0;
(4)  \textit{Piece-Wise Laplace Sampling}, due to model parameters changing much faster at earlier epochs of each learning schedule, sampling ${K}$ epochs ranging from $1$ to $T_{train}$ with multiple Laplace distributions centering at 0 and other learning schedule epochs (150 and 250 in our case). Since the searching difficulty grows exponentially with the size of $K$, another benefit of small candidate size is that it improves the searching efficiency. 


\section{Experiments}
\begin{table*}[!t]
	\centering
	\scriptsize
	\resizebox{\linewidth}{!}{%
		\begin{tabular}{cccccccccccccccc}
			\toprule
			\textbf{Dataset} & \textbf{Model} & \multicolumn{2}{c}{\textbf{Weight Decay}} &
			\multicolumn{2}{c}{\textbf{Brier Loss}} & \multicolumn{2}{c}{\textbf{MMCE}} &
			\multicolumn{2}{c}{\textbf{Label Smoothing}} & \multicolumn{2}{c}{\textbf{FL-3}} &
			\multicolumn{2}{c}{\textbf{FLSD-53}} &
			\multicolumn{2}{c}{\textbf{PCS}} \\
			\textbf{} & \textbf{} &
			\multicolumn{2}{c}{\cite{guo2017calibration}} &
			\multicolumn{2}{c}{\cite{brier1950verification}} & \multicolumn{2}{c}{\cite{kumar2018trainable}} &
			\multicolumn{2}{c}{\cite{szegedy2016rethinking}} & \multicolumn{2}{c}{\cite{mukhoti2020calibrating}} &
			\multicolumn{2}{c}{\cite{mukhoti2020calibrating}} &
			\multicolumn{2}{c}{Ours} \\
			&& Pre T & Post T & Pre T & Post T & Pre T & Post T & Pre T & Post T & Pre T & Post T & Pre T & Post T & Pre T & Post T \\
			\midrule
			\multirow{4}{*}{CIFAR-10/SVHN} & ResNet-50&94.32&94.56&93.59&93.72&85.17&64.75&78.88&78.89&88.28&88.42&92.48&92.79&\textbf{98.12}&\textbf{97.04}\\
		    & ResNet-110&61.71&59.66&94.80&95.13&85.31&85.39&68.68&68.68&96.74&\textbf{96.92}&90.83&90.97&\textbf{96.77}&96.77\\   
			& Wide-ResNet-26-10&96.82&97.62&94.51&94.51&97.35&97.95&84.63&84.66&98.19&98.05&\textbf{98.29}&\textbf{98.20}&97.55&97.84\\
			& DenseNet-121&84.43&81.57&94.65&94.66&85.88&84.87&78.79&78.94&89.48&89.42&89.59&89.59&\textbf{96.72}&\textbf{96.72}\\
			\midrule
			\multirow{4}{*}{CIFAR-10/CIFAR-10-C} & ResNet-50&86.23&86.03&\textbf{90.21}&\textbf{90.13}&89.97&90.11&72.01&72.02&89.44&89.56&89.45&89.56&89.73&89.79\\								
			& ResNet-110&77.53&75.16&84.09&83.86&71.96&70.02&72.17&72.18&82.27&82.18&85.05&84.70&\textbf{88.1}&\textbf{88.27}\\
			& Wide-ResNet-26-10&81.06&80.68&85.03&85.03&82.17&81.72&71.10&71.16&82.17&81.86&87.05&87.30&\textbf{89.62}&\textbf{89.95}\\
			& DenseNet-121&87.61&86.41&87.38&87.38&84.9&84.88&73.67&73.8&87.12&87.53&89.47&89.47&\textbf{89.52}&\textbf{89.52}\\
			\bottomrule
		\end{tabular}%
	}
	\caption{\textbf{Robustness on Dataset Shift}.\quad AUROC $(\%)$, being the higher the better, is evaluated for different methods with models shifting from CIFAR-10 (in-distribution) to SVHN and CIFAR-10-C as the OoD datasets. }
	\label{table:auroc_tab1}
\end{table*}
\subsection{Experimental Settings}
\noindent
\textbf{Datasets} \quad
We conduct experiments on various datasets, including CIFAR-10/100~\cite{krizhevsky2009learning} and Tiny-ImageNet~\cite{deng2009imagenet} to evaluate the calibration performance. We also include the robustness evaluation on Out-of-Distribution (OoD) datasets, including SVHN~\cite{goodfellow2013multi} and CIFAR-10-C~\cite{hendrycks2018benchmarking}.

\noindent
\textbf{Baselines} \quad
To verify the effectiveness of our proposed algorithm, we include different networks for evaluation, including ResNet-50, ResNet-110~\cite{he2016deep}, Wide-ResNet-26-10~\cite{zagoruyko2016wide} and DenseNet-121~\cite{huang2017densely}, and compare with various approaches, including training with weight decay at $5\times 10^{-4}$ (we find that weight decay at $5\times 10^{-4}$ performs the best among multiple values), Brier Loss~\cite{brier1950verification}, MMCE loss~\cite{kumar2018trainable}, Label smoothing~\cite{szegedy2016rethinking} with a smoothing factor $\alpha_{LS}=0.05$, focal loss~\cite{mukhoti2020calibrating} with regularisation parameter $\gamma_{focal}=3$, and scheduled focal loss FLSD-53~\cite{mukhoti2020calibrating} which 
uses $\gamma_{focal}=5$ for $\hat{p}\in[0,0.2)$ and $\gamma_{focal}=3$ for $\hat{p}\in[0.2,1)$.

\noindent
\textbf{Other Calibration Metrics} \quad
Recent works~\cite{nixon2019measuring,kumar2019verified,roelofs2022mitigating,gupta2020calibration} point out the defects of ECE. To evaluate our method comprehensively, we evaluate our method on three additional calibration metrics, i.e., MCE, Adaptive-ECE~\cite{ding2020revisiting} and classwise-ECE~\cite{kull2019beyond} along with ECE. We also measure PCS with reliability plots in supplementary.

\noindent
\textbf{Training Setup} \quad
For training on CIFAR-10/100, we set $T_{train}=350$. The learning rate is set to $0.1$ for epoch $0$ to $150$, $0.01$ for $150$ to $250$, and $0.001$ for $250$ until the end of training.
For training on Tiny-ImageNet, we set $T_{train}=100$. We follow the same training and validation set spilt setting as~\cite{mukhoti2020calibrating}. The learning rate is set to $0.1$ for epoch $0$ to $40$, $0.01$ for epoch $40$ to $60$, and $0.001$ for $60$ until the end of training. 
The fine-tuning learning rate is set to $10^{-4}$ for CIFAR-10, $5\times 10^{-4}$ for CIFAR-100, and  $10^{-3}$ for Tiny-ImageNet. The searching process is performed with $T_{se}=100$ steps. The population size is $\mc{S}=100$. 
Experiments are conducted with ResNet-50 on CIFAR-10 if there is no other specification. All networks are optimized using the SGD optimizer with a weight decay at $5\times 10^{-4}$ and a momentum of 0.9. The training batch size is set to 128. 
All experiments are conducted on a single Tesla V-100 GPU with all random seeds set to 1. Our code and results of comparison method are based on the public code and the pre-trained weight provided by~\cite{mukhoti2020calibrating}.

\noindent
\textbf{Temperature Scaling} \quad
Following the setting in the prior work~\cite{mukhoti2020calibrating}, the temperature parameter $\tau$ is optimized by grid searching with $\tau\in[0,0.1,0.2,\dots,10]$ on the validation set and finding the one with the best post-temperature-scaling ECE, which is also applied on the additional calibration metrics.

\subsection{Calibration Performance}
We report ECE$(\%)$ (computed using 15 bins) along with optimal temperatures in Table~\ref{table:ece_tab1}. PCS achieves the state-of-the-art ECE across all models and datasets and outperforms previous works by large margins, especially pre-temperature-scaling results. More specifically, most PCS pre-temperature-scaling results have already substantially exceeded the post-temperature-scaling results of previous works. The result of ResNet-110 on CIFAR-100 achieves the best calibration performance compared to previous works, with a $7\%$ decrease in ECE. For comparison approaches, the model trained with focal loss is broadly better-calibrated than other methods. However, it fails in some cases such as the evaluation on Tiny-ImageNet. In addition, the scheduled $\gamma_{focal}$ trick does not always work better than fixed $\gamma_{focal}$ and it is hard to ascertain which is the better between FL-3 and FL53. The MMCE auxiliary loss performs worst before temperature scaling. Another notable point of PCS is that multiple results such as those with ResNet-110 on CIFAR-10/100 achieve the innately calibrated model (T=1.0), which means that the PC itself has already yielded a well-calibrated model without temperature scaling. 

We also evaluate PCS on other widely-accepted metrics including Adaptive ECE, Classwise-ECE and test set error. PCS also achieves the state-of-the-art calibration results on almost all cases. The test set error on Tiny-ImageNet shows a $8.52\%$ decrease from $49.81\%$ to $41.29\%$, which is mainly because of PCS performing as an early stopping trick.

\subsection{Robustness on Out-of-Distribution(OoD) Datasets}
A well-calibrated model helps improve the model robustness on OoD datasets~\cite{thulasidasan2019mixup}. However, temperature scaling is known to be fragile under dataset distribution shift~\cite{ovadia2019can}. PCS form innately calibrated models and thus perform well on OoD datasets. We utilize AUROC (the higher the better) to evaluate the robustness under dataset shift. Table~\ref{table:auroc_tab1} shows the AUROC $(\%)$ computed for models trained on CIFAR-10 and tested on the OoD datasets SVHN and CIFAR-10-C. Our method achieves competitive results on almost all cases. The results after temperature scaling tend to drop generally, and approaches yielding better pre-temperature-scaling ECE have better robustness on OoD datasets. Although focal loss works well on calibration, it fails under dataset shift. Our PCS with ResNet-50 on CIFAR-10 achieves a $4\%$ increase compared to previous methods.

\begin{figure*}[t]
	\centering
	\includegraphics[width=\linewidth]{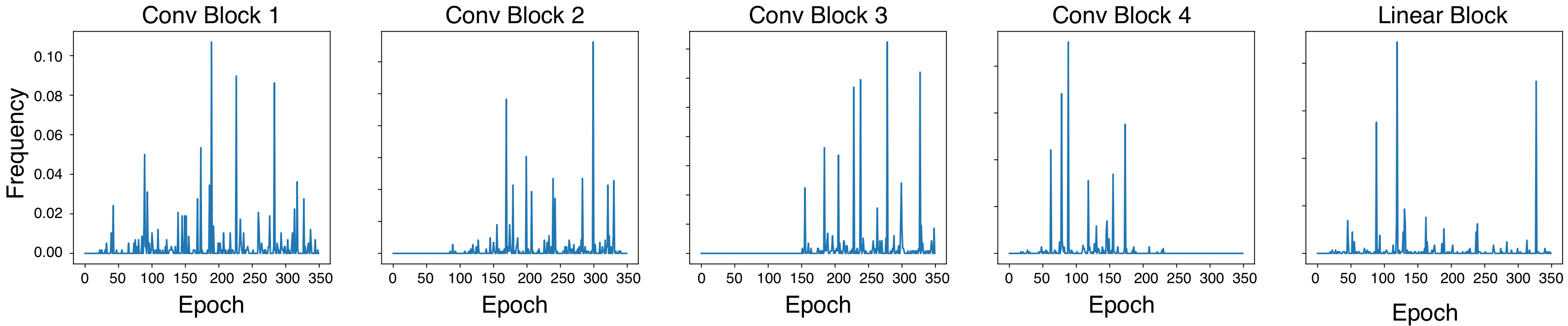}
	\caption{\textbf{Searching Results}.\quad
 The sub-figures show the predecessor choice frequency of different blocks in ResNet-50. The search results are filtered by cross-entropy loss (being less than 0.2) on the test set.
	}
 
	\label{fig:combination_stat}
\end{figure*}
\subsection{Searching Results}
We visualize the searching results of ResNet-50 on CIFAR-10 in Figure~\ref{fig:combination_stat}. The searching results are obtained based on well-fitting (test set loss $<$ 2) PCs. The result shows that the last convolutional block (Conv Block 4) tends to choose predecessors in the first half of training (epoch $50$ to $180$) while Conv Block 2 and Conv Block 3 prefer predecessors in the second half (epoch $150$ to $350$). This observation might indicate that the later Conv Blocks tend to overfit earlier than the former ones. The first Conv Block and the final linear block show no particular preference to certain predecessors. 

This result is consistent with the evaluation of overfitting of individual blocks in Figure~\ref{fig:motivation}. The Conv Block 1 suffers from little overfitting throughout training and thus has no preference to certain predecessors, while the middle blocks (Conv Block 2, 3, 4) prefer predecessors with a low validation loss as shown in Figure~\ref{fig:motivation}. 
We visualize the searching results of other models and observe the similar pattern. We also test our algorithm on other networks such as ViT~\cite{dosovitskiy2020image} and MLP-mixer~\cite{tolstikhin2021mlp}, which show the similar calibration effect.
\subsection{Ablation Study}
\noindent
\textbf{Comparison with Other Searching Methods} \quad
In Table~\ref{table:searching_methods}, our method is compared with different searching methods as well as early stopping methods. When early stopping on the model as a whole, it is hard to ensure a low error and good calibration performance at the same time. Early stopping on loss and ECE shows a large performance drop on classification error. The random search is conducted 5 times to make the results stable and achieve a relatively high performance on classification error but not ECE. We also compare multiple objectives in Eq.~(\ref{eq:objective3}) with the performance of a single objective optimization on validation NLL loss, i.e.,$\operatorname*{min}_A \hat{NLL}$. Searching on ECE or error individually could lead to extremely unbalanced results. PCS achieves a better result on both test set error and ECE. The hyperparameter $\lambda$ is tuned on models and datasets.
\begin{table}[!t]
	\centering
	\scriptsize
	\begin{tabular}{ccccc}
		\toprule
		\textbf{Methods} & \textbf{ERR} & \textbf{ECE} & \textbf{ECE(T)}\\
		\midrule
		\textbf{Early Stopping (Loss)} & 6.58 & 2.32 & 0.56(1.4)\\
		\textbf{Early Stopping (Error)}& 4.89 & 4.03 & 1.42(2.4)\\
		\textbf{Early Stopping (ECE)}& 16.92 & 1.71 & 1.09(1.2)\\
		\textbf{Random Search} & 5.04 & 1.13 & 0.84(1.1)\\
		\textbf{Search on Loss}& 5.43 & 1.1 & 0.37(1.10)\\
		\textbf{PCS ($\lambda$=10)}& 5.31 & 0.86 & 0.73(1.10)\\
		\textbf{PCS ($\lambda$=25)}& 5.1 & 0.75 & 0.37(1.10)\\
		\textbf{PCS ($\lambda$=50)}& 5.25 & 0.58 & 0.58(1.0)\\
		\textbf{PCS ($\lambda$=100)}& 5.29 & 0.70 & 0.44(1.10)\\
		\bottomrule
	\end{tabular}
	\caption{\textbf{Comparison of Searching Methods}.\quad Random search is conducted 5 times with ResNet-50 on CIFAR-10. All searching results are selected by the lowest testing loss.}
	\label{table:searching_methods}
\end{table}

\begin{figure}[t]
	\centering
	\includegraphics[width=1\linewidth]{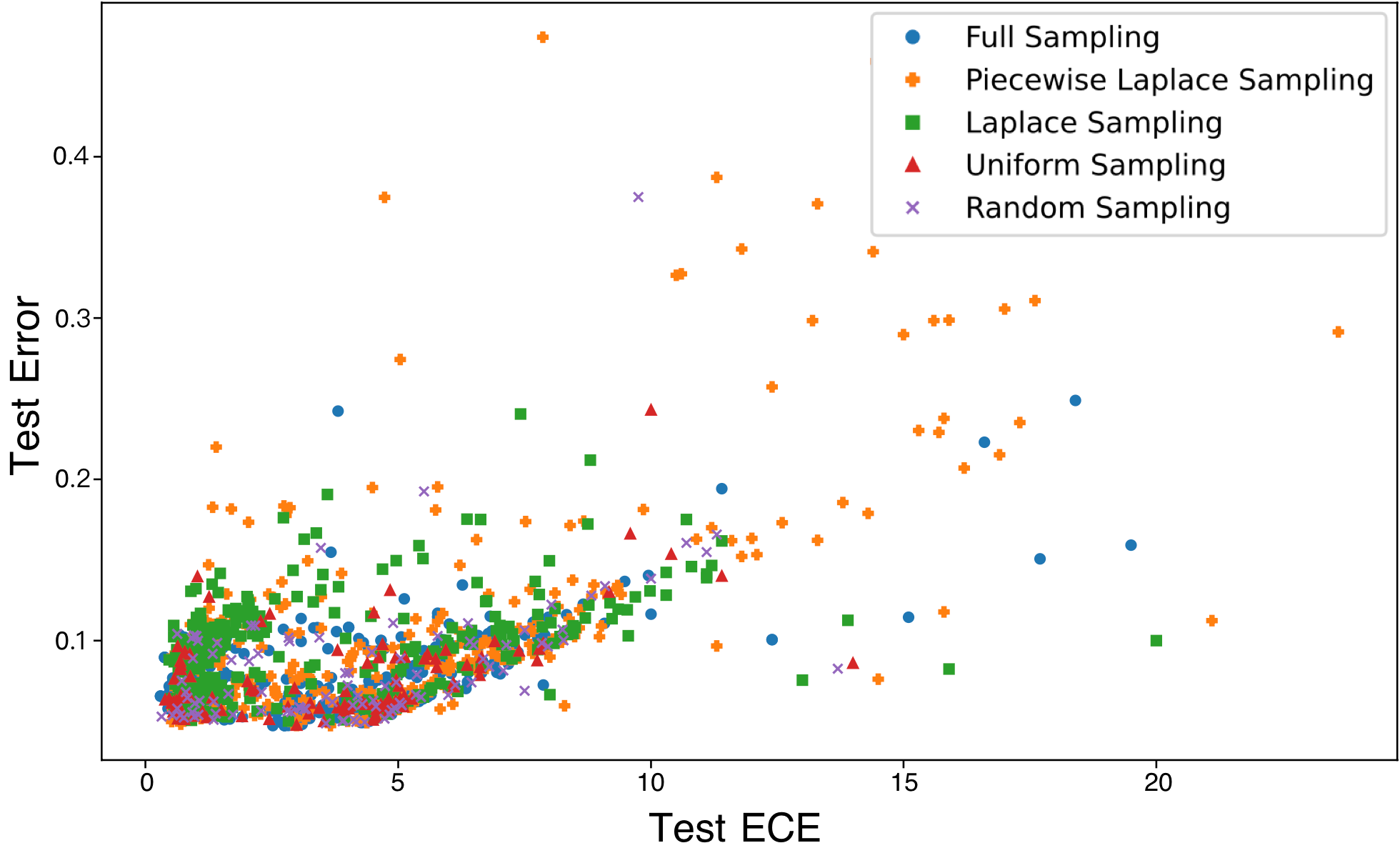}
	\caption{\textbf{Comparison of Sampling Strategy}.\quad
		All experiments are conduct 5 runs with ResNet-50 on CIFAR-10. Searching step $T_{se}$ is set to $100$ and produces 500 searching results for each strategy. Metrics along the x-axis and y-axis are the lower the better.
	}
	\label{fig:sampling_strategy}
\end{figure}

\noindent
\textbf{Weight Sampling Strategy} \quad
To compare sampling strategies discussed in the previous section, a scatter plot in Figure~\ref{fig:sampling_strategy} shows the searching results. Note that the ``Full Sampling'' indicates searching on all possible predecessors without sampling, i.e., $K=T_{train}$. We use the same $K=50$ for all sampling strategies. From Figure~\ref{fig:sampling_strategy}, we observe little difference between different sampling strategies, which indicates that we save storage space with little loss of performance. Thus, we use the random sampling strategy throughout the paper due to its simplicity.

\noindent
\textbf{Warm-up Population} \quad
A larger population indicates more searching time. To find a balance between searching time and performance, we compare the searching results with different population sizes. We use the number of well-fitting results (test loss under 0.2) to measure the searching performance. All experimental results are averaged over 5 runs. According to Table~\ref{table:population}, the larger the population size, the more well-fitting results can be found since estimator $\psi$ can have better prior knowledge on error and ECE landscape. However, we use a smaller population size as long as the searching provides satisfactory results. 
\begin{table}[!h]
	\centering
	\scriptsize
	\begin{tabular}{ccccc}
		\toprule
		\textbf{Population Size} & \textbf{64} & \textbf{100} & \textbf{200} & \textbf{500}\\
		\midrule
		\textbf{GPU Hours} & 3.1 & 3.7 & 5.5 & 10.5 \\
		\textbf{Well-fitting Results}& 14.8 & 16 & 17.8 & 25.8 \\
		\bottomrule
	\end{tabular}
	\caption{\textbf{Warm-up Population}.\quad Well-fitting results indicate the number of searching results that achieve a testing loss under 0.2, being the higher the better. All results are reported as an average of 5 runs with ResNet-50 on CIFAR-10.}
	\label{table:population}
\end{table}

\section{Conclusion}
In this paper, we address a common problem, the mis-calibration in modern neural networks. We observe that different blocks in a network have different overfitting patterns. Our proposed predecessor combination search, as a regularization method, is very effective for calibrating models and can also be potentially applied to other tasks such as learning with noisy labels and improving model robustness. 



\section*{Acknowledgments}
This work was supported in part by the Australian Research Council under Project DP210101859 and the University of Sydney Research Accelerator (SOAR) Prize. The authors acknowledge the use of the National Computational Infrastructure (NCI) which is supported by the Australian Government, and accessed through the NCI Adapter Scheme and Sydney Informatics Hub HPC Allocation Scheme.

\bibliographystyle{named}
\bibliography{ijcai23}

\end{document}